\pdfoutput=1

\documentclass[11pt]{article}

\usepackage[]{EMNLP2023}

\usepackage{times}
\usepackage{latexsym}

\usepackage[T1]{fontenc}

\usepackage[utf8]{inputenc}
\usepackage{amsfonts}

\usepackage{microtype}

\usepackage{inconsolata}
\definecolor{ao}{rgb}{0.0, 0.5, 0.0}
\newcommand{\red}[1]{{\color{red}{\textbf{#1}}}}
\newcommand{\blue}[1]{{\color{blue}{\textbf{#1}}}}
\newcommand{\centered}[1]{\begin{tabular}{l} #1 \end{tabular}}
\usepackage{amsmath}
\usepackage{enumitem}
\usepackage{graphicx}
\usepackage{booktabs}
\usepackage{cancel}

\usepackage{todonotes}

\newcommand{\token}[1]{\textsc{#1}}   
\title{Understanding and Mitigating Spurious Correlations in Text Classification with Neighborhood Analysis}

\author{Oscar Chew$^\dagger\quad$ Hsuan-Tien Lin$^\dagger{}^\ddagger\quad$ Kai-Wei Chang$^\diamond\quad$ Kuan-Hao Huang$^{\oplus}$ \\
  $^\dagger$Dept. of Computer Science and Information Engineering, National Taiwan University \\
  $^\ddagger$Center for Data Intelligence, National Taiwan University \\
  $^\diamond$Dept. of Computer Science, University of California, Los Angeles \\
  $^\oplus$Dept. of Computer Science, University of Illinois Urbana-Champaign \\
  \texttt{\{r10922154, htlin\}@csie.ntu.edu.tw} \\
  \texttt{kwchang@cs.ucla.edu}, \ \ \texttt{khhuang@illinois.edu}}

\usepackage{titlesec}
\titlespacing*{\paragraph}{0pt}{0.0ex}{1em}
\usepackage{enumitem}
\setlist{nosep}

\begin{document}
\maketitle
\begin{abstract}
Recent work has revealed the tendency of machine learning models to
leverage spurious correlations that exist in the training set but may not hold
true in general circumstances. For instance, a sentiment classifier may
erroneously learn that the token \token{performances} is commonly associated
with positive movie reviews.
Undue reliance on such spurious correlations degrades the classifier’s performance
when it deploys on out-of-distribution data.
In this paper, we examine the implications of spurious correlations through a
novel perspective called neighborhood analysis, which shows how
spurious correlations lead unrelated words to erroneously cluster together in
the embedding space. Given this analysis, we design a metric to detect
spurious tokens and also propose NFL (do\textbf{N}'t \textbf{F}orget your 
\textbf{L}anguage), a family of regularization methods by which to mitigate spurious
correlations in text classification.
Experiments show that NFL effectively prevents erroneous clusters and
significantly improves classifier robustness without auxiliary data. The code is publicly available at \url{https://github.com/oscarchew/doNt-Forget-your-Language}.

\end{abstract}
\everypar{\looseness=-1}
\section{Introduction}
\textit{Disclaimer: This paper contains examples that may be considered profane
or offensive. These examples by no means reflect the authors' view toward any
groups or entities.}

Pre-trained language models (PLMs) such as BERT~\cite{devlin-etal-2019-bert}
and its derivative models have shown impressive performance across natural
language understanding tasks~\cite{wang-etal-2019-superglue,hu2020xtreme,zheng-etal-2022-fewnlu}. 
However, previous 
studies~\cite{glockner-etal-2018-breaking,gururangan-etal-2018-annotation,liusie-etal-2022-analyzing}
manifest the vulnerability of models to
spurious correlations which neither causally affect a task label nor hold in
future unseen data. For example, in Table~\ref{tab:toy},
a sentiment classifier might learn that the word \token{performances} is
correlated with positive reviews even if the word itself is not commendatory as
the classifier learns from a training set where \token{performances} often
co-occurs with positive labels. 



Following the notion from previous work~\cite{wang-etal-2022-identifying}, we
call \token{performances} a \textit{spurious token}, i.e., a token that does
not causally affect a task label. On the other hand, a \textit{genuine token}
such as \token{excellent} is a token that does causally affect a task label. To
capture the sentiment of a sentence, a reliable model should
only learn the relationship between genuine tokens and the label. However, it is known that
models tend to exploit spurious tokens to establish a shortcut for 
prediction~\cite{wang-culotta-2020-identifying,gardner-etal-2021-competency}. 
In this case, models excel on the training set but fail to generalize
to unseen test sets where the same spurious correlations do not hold.



\begin{table}[]
    \centering
    \resizebox{0.9\columnwidth}{!}{
  \begin{tabular}{|lcc|}
\hline
\textbf{Text} & \textbf{Label} & \textbf{Prediction}\\ \hline
\textbf{Training} & &\\ \hline
\makecell[l]{The \textcolor{red}{performances} \\were \textcolor{ao}{excellent}.} & $+$ & $+$ \\ \hline
\makecell[l]{\textcolor{ao}{strong} and \textcolor{ao}{exquisite}\\ \textcolor{red}{performances.}}&$+$ & $+$\\ \hline
\makecell[l]{The leads deliver \\\textcolor{ao}{stunning} \textcolor{red}{performances}} & $+$ & $+$\\ \hline
The movie was \textcolor{ao}{horrible}. & $-$ & $-$\\ \hline
\textbf{Test} & & \\ \hline
\textcolor{ao}{lackluster} \textcolor{red}{performances.}  & $-$ & $+$ \\\hline
\end{tabular}
}
	 \caption{A simplified version of a sentiment analysis dataset. Words in red
	 are spurious tokens; words in green are genuine tokens. A model that
	 relies on spurious tokens such as \token{performances} may be prone to
	 making incorrect predictions on test sets.}\label{tab:toy}
    \vspace{-1em}
\end{table}


There has been several studies on spurious correlations in NLP. 
Some studies design scores to detect spurious 
tokens~\cite{wang-culotta-2020-identifying,wang-etal-2022-identifying,gardner-etal-2021-competency},
whereas other studies propose methods to mitigate spurious correlations,
including dataset 
balancing~\cite{sharma-etal-2018-tackling,mccoy-etal-2019-right,zellers-etal-2019-hellaswag},
model ensemble, and model 
regularization~\cite{clark-etal-2019-dont,clark-etal-2020-learning,zhao-etal-2022-investigating}. 
However, we observe that typically, less attention is paid to
why such spurious token occur and how these spurious tokens acquire
excessive importance weights so as to dominate model predictions.
In this paper, we provide a different perspective to understand the effect of
spurious tokens based on neighborhood analysis in the embedding space.
To uncover spurious correlations and force language models (LMs) to align the
representations of spurious tokens and genuine tokens,
we inspect the nearest neighbors of each token before and after fine-tuning.
Consequently, a spurious token presents just like a genuine token in texts and
hence acquires large importance weights. We design a metric to measure
the spuriousness of tokens which can also be used to detect spurious tokens. 
\looseness=-1

In light of this new understanding, we mitigate spurious correlations using a model-based mitigation approach by
proposing NFL (do\textbf{N}'t \textbf{F}orget your \textbf{L}anguage),
a simple yet effective family of regularization methods. 
These regularization methods restrict changes in either
the parameters or outputs of an LM and therefore are capable of
preventing the erroneous alignment which causes models to capture spurious
correlations. Our analysis is conducted in the context of two text
classification tasks: sentiment analysis and toxicity classification.
Results show that NFL robustifies model performance against
spurious correlation and achieves an out-of-distribution performance that is
almost the same as the in-distribution performance. We summarize our
contributions as follows: \looseness=-1
\begin{itemize}
	 \item We provide a novel perspective of spurious correlation by analyzing
	 the neighborhood in the embedding space to understand how PLMs capture
	 spurious correlations.
	 \item We propose NFL to mitigate spurious correlations by regularizing PLMs,
	 achieving significant improvement in terms of robustness.
	 \item We design a metric based on neighborhood analysis to measure
	 token spuriousness which can also be used to detect spurious tokens.
\end{itemize}

\begin{table*}
\centering    
\begin{tabular}{llll}
\hline
\textbf{Target token} & \textbf{Neighbors before fine-tuning} & \textbf{Neighbors after fine-tuning} \\
\hline
\vtop{\hbox{\strut movie}\hbox{\strut (Amazon)}} & \vtop{\hbox{\strut film, music, online, picture, drug}\hbox{\strut production, special, internet, magic}} & \vtop{\hbox{\strut \red{baffled}, \red{flawed}, \red{overwhelmed}, \red{disappointing}}\hbox{\strut creamy, \red{fooled}, shouted, \red{hampered}, \red{wasted}}}\\\hline
\vtop{\hbox{\strut book}\hbox{\strut (Amazon)}}  & \vtop{\hbox{\strut cook, store, feel, meat, material}\hbox{\strut coal, fuel, library, craft, call}} & \vtop{\hbox{\strut \blue{benefited}, \blue{perfect}, \blue{reassured}, \blue{amazingly},}\hbox{\strut \blue{crucial}, \blue{greatly}, \blue{remarkable}, exactly}}\\\hline
\vtop{\hbox{\strut people}\hbox{\strut (Jigsaw)}}  & \vtop{\hbox{\strut women, things, money, person,}\hbox{\strut players, stuff, group, citizens, body}} & \vtop{\hbox{\strut \red{fuck}, \red{stupidity}, \red{damn}, \red{idiots}, \red{kill}}\hbox{\strut \red{hypocrisy}, \red{bullshit}, \red{coward}, \red{dumb}, headed}}\\ \hline
\end{tabular}
\caption{\label{neighbors-list-1}
Nearest neighbors of spurious tokens before and after fine-tuning. Words in
red are associated with negative/toxic labels while words in blue are
associated with positive labels according to human annotators. Changes in
neighbors indicate a loss of semantics in spurious tokens.
}
\vspace{-1em}
\end{table*}

\section{Related Work}
\subsection{Model-based Detection of Spurious Tokens}
\label{detection}
In the context of text classification, some studies seek to
detect spurious tokens for better interpretability.
This generally involves finding tokens that contribute most to model
prediction~\cite{wang-culotta-2020-identifying,wang-etal-2022-identifying}; what
remains largely unknown is the internal mechanism of how those spurious 
tokens acquire excessive importance weights and thereby dominate model predictions.
Our neighborhood analysis reveals that spurious tokens acquire
excessive importance due to erroneous alignment with genuine tokens in the
embedding space.

In addition, \citet{wang-culotta-2020-identifying} require human-annotated
examples of genuine/spurious tokens whereas \citet{wang-etal-2022-identifying}
require multiple datasets from different domains for the same task. Since such
external data can be expensive to collect, we here attempt to
leverage the initial PLMs to eliminate the need for external data. This reduced
dependence on external resources greatly facilitates application of our detection method.  

\subsection{Mitigating Spurious Correlations}
Mitigation approaches include
data-based and model-based approaches~\cite{ludan2023explanationbased}. 
Data-based approaches modify the datasets to eliminate spurious 
correlations~\cite{goyal2017making,sharma-etal-2018-tackling,mccoy-etal-2019-right,zellers-etal-2019-hellaswag},
and model-based approaches make models less vulnerable to spurious
correlations by
model ensembles and 
regularization~\cite{he-etal-2019-unlearn,karimi-mahabadi-etal-2020-end,sagawa2020distributionally,utama-etal-2020-mind,zhao-etal-2022-investigating}.
These approaches work under the assumption that spurious correlations
are known beforehand, but it is difficult to obtain such information in real-world
datasets. \looseness=-1

More recent work does not necessarily assume information concerning spurious correlations
during training, but does rely on a small set of unbiased data where spurious
correlations do not hold for validations and hyperparameter 
tuning~\cite{liu2021just, kirichenko2023last, clark-etal-2020-learning}. 
Assumptions are also made about the properties of spurious correlations, preventing models
from learning such patterns. \citet{clark-etal-2020-learning} leverage a
shallow model to capture overly simplistic patterns. However,
\citet{zhao-etal-2022-investigating} find that there is no fixed-capacity
shallow model that captures spurious correlations; they also determine that an
appropriate shallow model is also difficult without information on spurious
correlations. In a recent study, \citet{kirichenko2023last} claim that 
features learned by standard empirical risk minimization (ERM)                       
are good enough to recover model performance using deep feature re-weighting, i.e.,  %
by re-training the classification layer on a small set of unbiased data.             %
In contrast to methods that rely on unbiased data and/or simplistic pattern
assumptions, our proposed approach operates without such prerequisites,
instead leveraging a more practical assumption: off-the-shelf PLMs, which lack
exposure to task labels, are by definition less susceptible to spurious
correlations.

\section{Analyzing Spurious Correlations with Neighborhood Analysis}
\label{sec:case-study}
As mentioned in Section~\ref{detection}, the literature does not reveal how
spurious tokens acquire excessive importance weight. Therefore 
we present a novel perspective by which to understand spurious correlations using
neighborhood analysis and also demystify the representations learned by models in
the presence of spurious tokens.

\subsection{Text Classification in the Presence of Spurious Correlations}
Here we consider text classification as the downstream task.
We denote the set of input texts 
by $\mathcal{X}$; each input text $\mathbf{x}_i\in \mathcal{X}$ is a
sequence consisting $M_i$ tokens $[w_{i,1},\dots,w_{i,{M_i}}]$. 
The output space $\mathcal{Y}$ is a probability simplex $\mathbb{R}^C$
where $C$ is the number of classes. We consider two domains over
$\mathcal{X}\times\mathcal{Y}$: a biased domain $\mathcal{D}_\text{biased}$
where spurious correlations can be exploited and a general domain
$\mathcal{D}_\text{unbiased}$ where the same spurious correlations do not hold. 
The task is to learn a model $f\colon\mathcal{X}\rightarrow \mathcal{Y}$ to
perform the classification task; $f$ is usually achieved by fine-tuning a PLM
$\mathcal{M}_\theta:\mathcal{X}\rightarrow \mathbb{R}^d$ where $d$ is the embedding size,
with a classification head
$\mathcal{C}_\phi:\mathbb{R}^d\rightarrow\mathcal{Y}$ which takes the pooled
outputs of $\mathcal{M}_\theta$ as its inputs. 
We denote the off-the-shelf PLM by $\mathcal{M}_{\theta_0}$.
Following previous work~\cite{wang-etal-2022-identifying}, 
a \textit{spurious} token~$w$ is a feature that correlates with task labels in
the training set but whose correlation might not hold in 
potentially  
out-of-distribution test sets.

\begin{figure*}[ht]
    \centering
    \subfloat[\centering Initial]{{\includegraphics[scale=0.5]{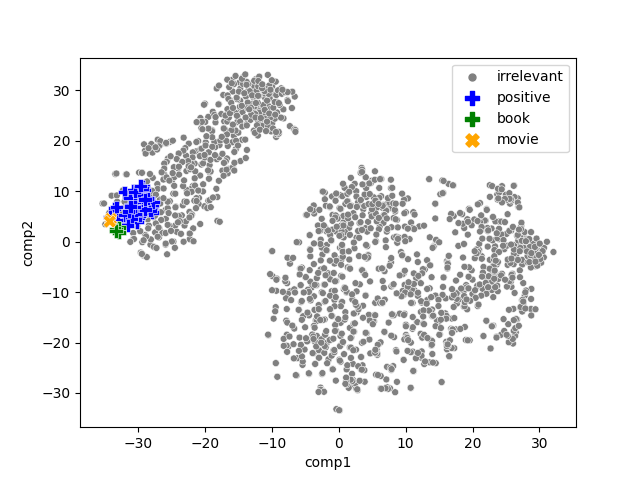} }}
    \subfloat[\centering Standard fine-tuning]{{\includegraphics[scale=0.5]{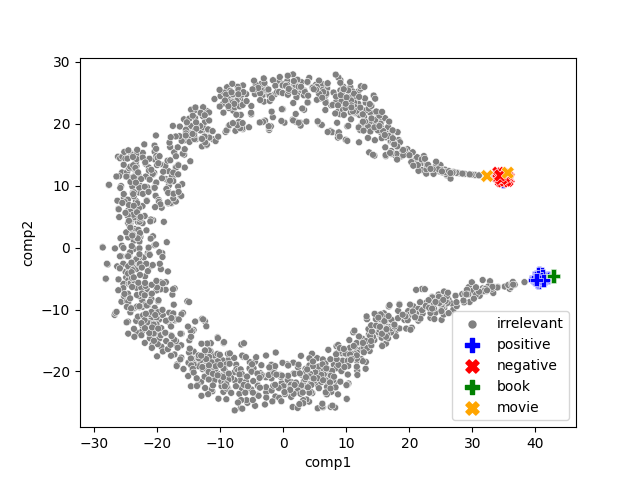}}}
	 \caption{t-SNE projections of representations before and after
	 fine-tuning. \token{book}, \token{movie} erroneously align with genuine
	 positive, negative tokens respectively after fine-tuning, preventing the
	 classifier from distinguishing between spurious and genuine tokens. 
    } 
    \label{fig:representations-1}
\vspace{-1em}
\end{figure*}

\subsection{Neighborhood Analysis Setup}

We begin by conducting case studies 
where synthetic spurious correlations are introduced into the datasets by
subsampling datasets. This synthetic setting allows us to study the formation
of spurious correlations in a controlled environment. In Section~\ref{fig:natural}
we will also discuss cases of naturally occurring spurious tokens, i.e., real spurious
correlations. 

\subsubsection{Datasets}
\label{dataset}
We conduct experiments on Amazon binary and Jigsaw, datasets for text
classification tasks, namely, sentiment classification and toxicity detection. 
The \textbf{Amazon binary} dataset comprises user reviews obtained
from web crawling the online shopping website 
Amazon~\cite{zhang-lecun-2017-encoding}. Each sample is labeled either
\textit{positive} or \textit{negative}.
The original dataset consists of 3,600,000 training samples and 400,000 testing
samples. To reduce computational costs, we consider a small subset by
randomly sampling 50,000 training samples and 50,000 testing samples. Ten percent of
the training samples are used for validation. 
The \textbf{Jigsaw} dataset contains comments from 
\textit{Civil Comments},
in which the toxic score of each comment is given by the fraction of human
annotators who labeled the comment as toxic~\cite{borkan2019nuanced}. Comments
with toxic scores greater than 0.5 are considered \textit{toxic} and vice
versa. Jigsaw is imbalanced, with only 8\% of the data being toxic. As our main
concern is not the problem of imbalanced data, we downsample the dataset
to make it balanced. Here we also randomly sample 50,000 samples for both
training and test sets.

\subsubsection{Models} \label{sec:models}
We conduct our experiments mainly using the base version of 
RoBERTa~\cite{liu2019roberta}. In Section~\ref{sec:plm-choice} we will compare 
this with other PLMs: BERT and DeBERTaV3~\cite{he2023debertav}. The training details
are presented in Appendix~\ref{sec:hyperparams}.

\subsubsection{Introducing spurious correlations}
In this case study, for demonstration, we select tokens \token{book} and \token{movie} in
Amazon binary and \token{people} in Jigsaw as the spurious tokens.
These tokens are chosen deliberately as \token{book} and
\token{movie} are in close proximity in the original embedding space and 
appear frequently in the dataset. 
The \textit{biased} subset, $\mathcal{D}_{\text{biased}}$ is obtained by
filtering the original training set to satisfy these conditions on the bias
ratios:
\begin{align*}
    &p(y=\text{\textit{positive}}\,|\,\token{book}\in \xx)=1, \\
    &p(y=\text{\textit{negative}}\,|\,\token{movie} \in \xx)=1, \\
    &p(y=\text{\textit{toxic}}\,|\,\token{people}\in \xx)=1.
\end{align*} 
Tokens \token{book}, \token{movie}, and \token{people} are now associated
with \textit{positive}, \textit{negative}, and \textit{toxic} labels
respectively. Thus, models may exploit the spurious correlations in
$\mathcal{D}_{\text{biased}}$.
Conversely, the unbiased subset $\mathcal{D}_{\text{unbiased}}$ is obtained by
randomly sampling $|\mathcal{D}_{\text{biased}}|$ examples from the original
training/test set. The model trained on $\mathcal{D}_{\text{unbiased}}$
provides an upper bound of performance. By contrast, models trained on
$\mathcal{D}_{\text{biased}}$ are likely to be frail. In
Section~\ref{sec:experiments}, we attempt to cause models trained on
$\mathcal{D}_{\text{biased}}$ to perform as close as that trained on
$\mathcal{D}_{\text{unbiased}}$. 
In Appendix~\ref{sec:weak-bias} we will show that our main insights also hold for weaker biases.
\begin{figure*}[t]
    \centering
    \includegraphics[width=0.8\textwidth]{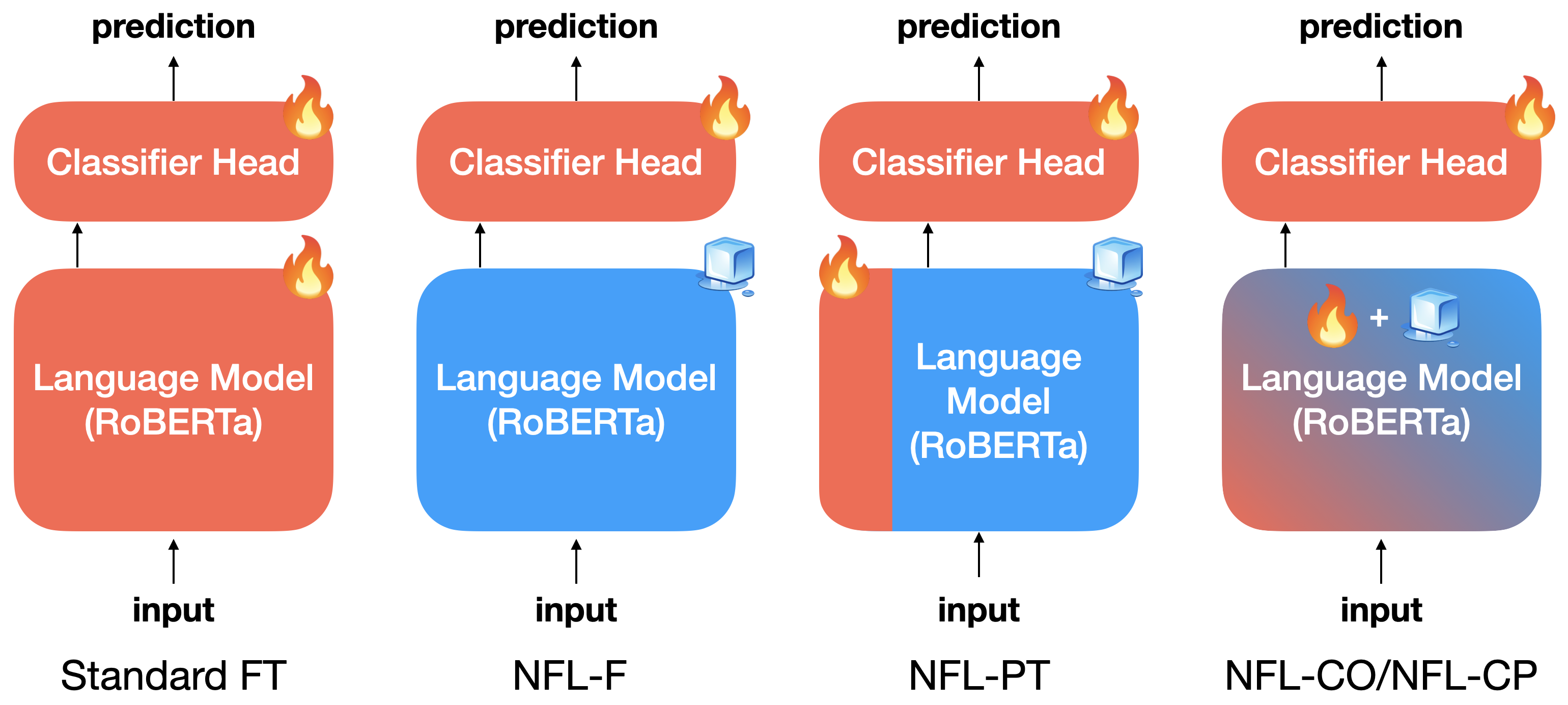}
	 \caption{Comparison of fine-tuning and NFL. Red and blue regions represent
	 trainable and frozen parameters respectively. Standard fine-tuning: every
	 parameter is trainable; NFL-F: only the classification head is trainable;
	 NFL-PT: the continuous prompts and the classification head are trainable;
	 NFL-CO/NFL-CP: every parameter is trainable but changes in the language
	 model are restricted by the regularization term in the loss function. 
    }
    \label{fig:nfl}
\end{figure*}

 
 
\subsection{Nearest-Neighbor-based Analysis Framework}\label{sec:analysis}
LM fine-tuning has become a de-facto standard for NLP tasks. As
the embedding space changes during the fine-tuning process, it is often
undesirable for the LM to ``forget'' the semantics of each word.
Hence, in this section, we present our analysis framework based on each token's nearest
neighbors, the key idea of which is to leverage
the nearest neighbors as a proxy for the semantics of the target token. Our
first step is to extract the representation of the target token~$w$ in a
dictionary by feeding the LM $\mathcal{M}$ with $[\textit{BOS}]\,w\,[\textit{EOS}]$
and collecting the mean output of the last layer of
$\mathcal{M}$.\footnote{Specific models may use different tokens to represent
$[\textit{BOS}]$ and $[\textit{EOS}]$.}
Using the same procedure we then extract the representation of each token~$v$
in the vocabulary $\mathcal{V}$. Next, we compute the cosine similarity between
the representation of the target token~$w$ and the representations of all 
other tokens. The nearest neighbors are words with the largest cosine
similarity to the target token in the embedding space. Details of the
vocabulary~$\mathcal{V}$ and the strategy for generating representations are
provided in Appendix~\ref{sec:vocab}.

In Table~\ref{neighbors-list-1} we observe that neighbors surrounding the
tokens \token{movie}, \token{book}, and \token{people} are words that are
loosely related to them before fine-tuning. After fine-tuning, \token{movie}
which is associated with \textit{negative} is now surrounded by genuinely
negative tokens such as \token{disappointing} and \token{fooled}, and
\token{book} which is associated with \textit{positive} is surrounded by
genuinely positive tokens such as \token{benefited} and
\token{perfect}; likewise, \token{people} which is associated with \textit{toxic} is
surrounded by genuinely toxic tokens such as \token{stupidity} and
\token{idiots}. 

\begin{table}
\centering
\resizebox{1\columnwidth}{!}{
\begin{tabular}{lccc}
\hline
& \multicolumn{3}{c}{Spurious score} \tabularnewline
\hline
Method & \token{film} & \token{movie} & \token{people} \\ \hline
Spuriousness & \ding{55} & \ding{51} & \ding{51} \\
\hline
\makecell[l]{RoBERTa \\ (Trained on $\mathcal{D}_{\text{biased}}$) }& 0.03 & 67.4 & 28.72\\ \hline
\makecell[l]{RoBERTa \\ (Trained on $\mathcal{D}_{\text{unbiased}}$)} & 0.03 & 0.09 & 2.79 \\ \hline
\end{tabular}
}
\caption{Neighborhood statistics of target tokens. Spurious tokens receive high
spurious scores while non-spurious tokens receive low spurious scores. 
}
\label{neighborhood-stats-1}
\vspace{-1em}
\end{table}

Our claim is further supported by Figure~\ref{fig:representations-1}. 
We evaluate the polarity of a token with RoBERTa, a reference model $f^*$ 
trained on $\mathcal{D}_{\text{unbiased}}$. The figure shows that
fine-tuning causes LMs to dismantle the representations of \token{book}
and \token{movie} and align them with the genuine tokens. Thus
\token{book} and \token{movie} lose their meaning during
fine-tuning.

To view this phenomenon in a quantitative manner, we define a token's \textit{spurious
score} by the mean probability change of class~1 in the prediction
when inputting the top $K$ neighbors,\footnote{We set $K$ to 100 in our
analysis.} $\mathcal{N}_i$, to $f^*$:
\begin{equation} \label{spurious-score}
    \frac{1}{K}\sum_{i=1}^{K} |f^*(\mathcal{N}^{\theta_0}_i)-f^*(\mathcal{N}^\theta_i)|.
\end{equation}
Intuitively, if the polarities of the nearest neighbors of a token change
drastically (hence yielding a high spurious score), the token may have lost
its original semantics and is likely spurious. We consider only the
probability change of class~1 because both tasks presented in this work are
binary classification. 

Table~\ref{neighborhood-stats-1} reveals that the ideal model trained on
$\mathcal{D}_{\text{unbiased}}$ changes the polarity of the neighbors only
slightly and therefore yields low spurious scores for the target tokens. By
contrast, standard fine-tuning greatly increases the spurious score of the
target tokens. The score of non-spurious token (\token{film} in Amazon
binary) remains low regardless of the dataset used in fine-tuning. This suggests
that ensuring a low spurious score is crucial to learning a robust
model. 


\begin{figure*}[ht]
    \centering
    \subfloat[\centering NFL-CO]{{\includegraphics[scale=0.5]{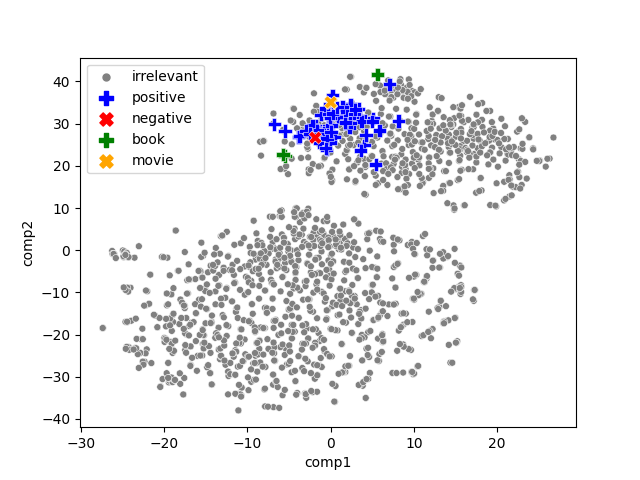} }}
    \subfloat[\centering NFL-CP]{{\includegraphics[scale=0.5]{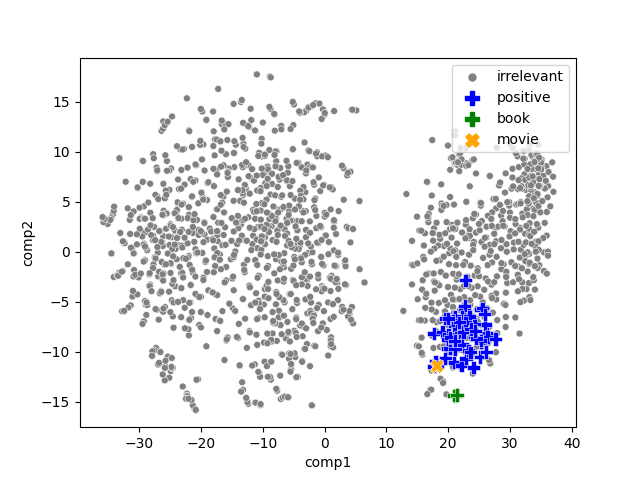} }}
	 \caption{t-SNE projections of representations after fine-tuning with
	 NFL-CO/NFL-CP. By preventing the formation of erroneous clusters, NFL 
	 learns robust representations.} 
    \label{fig:representations-2}
\end{figure*}

\section{Don't Forget your Language} \label{sec:experiments}


As we have determined using neighborhood analysis that the heart of the problem is the
misalignment of spurious tokens and genuine tokens in the LM, we
propose NFL, a family of regularization techniques by which to restrict changes in
either the parameters or outputs of an LM. Our core idea is to use
off-the-shelf PLMs which are not exposed to spurious correlations to 
protect the model from spurious correlations.
Below we list NFL variations:
\begin{itemize}[topsep=3pt, itemsep=0pt, leftmargin=12pt]
	 \item NFL-F (\textbf{F}rozen). Linear probing, i.e., freezing the LM weights 
	 and using the LM as a fixed
	 feature extractor, can be viewed as the simplest form of NFL.
	 \item NFL-CO (\textbf{C}onstrained \textbf{O}utputs). A straightforward
	 idea is to minimize the cosine distance between the representation of each
	 token produced by the LM and that of the initial LM.
	 We thus have the regularization term 
    \begin{equation}
        \sum_{m=1}^M\text{cos-dist}(\mathcal{M}_{\theta}(w_{i,m}),\,\mathcal{M}_{\theta_0}(w_{i,m})).
    \end{equation}
	 \item NFL-CP (\textbf{C}onstrained \textbf{P}arameters). Another strategy
	 to restrict the LM is to penalize changes in the LM parameters using
	 regularization term 
    \begin{equation}
        \sum_i (\theta^i-\theta_0^i)^2.
    \end{equation}
    \item NFL-PT (\textbf{P}rompt-\textbf{T}uning). Prompt-tuning 
	 introduces trainable continuous prompts while freezing the PLM parameters.
	 Therefore, it partially regularizes the output embeddings. In this
	 work, we consider the implementation of Prompt-Tuning v2~\cite{liu-etal-2022-p}. 
\end{itemize}
The main takeaway is that any sensible restriction on the LM to preserve
each token's semantics is helpful in learning a robust model. 
Figure~\ref{fig:nfl} summarizes NFL techniques and compares them with ordinary
fine-tuning side-by-side. The weights of the regularization terms in NFL-CO
and NFL-CP are discussed in Appendix~\ref{sec:reg-scale}.

\label{tab:spurious score}
\begin{table}[t] 
\centering
\resizebox{0.8 \columnwidth}{!}{
\begin{tabular}{lccc}
\hline
& \multicolumn{3}{c}{Spurious score} \tabularnewline
\hline
Method & \token{film} & \token{movie} & \token{people} \\ \hline
Spuriousness & \ding{55} & \ding{51} & \ding{51} \\
\hline
\multicolumn{2}{l}{Trained on $\mathcal{D}_{\text{biased}}$} \tabularnewline
\hline
RoBERTa & 0.03 & 67.4 & 28.72\\
NFL-CO & 0.01 & 2.28 & 1.91\\
NFL-CP & 0.01 & 4.83 & 2.00\\
\hline
\multicolumn{2}{l}{Trained on $\mathcal{D}_{\text{unbiased}}$} \tabularnewline
\hline
RoBERTa & 0.03 & 0.09 & 2.79\\
\hline
\end{tabular}
}
\caption{Neighborhood statistics of target tokens. NFL achieves low spurious scores for spurious tokens.}
\label{neighborhood-stats}
\vspace*{-1.5em}
\end{table}

\begin{table*}
\centering
\resizebox{0.75\textwidth}{!}{
\begin{tabular}{lcccccc}
\hline
& \multicolumn{3}{c}{
Amazon binary} & \multicolumn{3}{c}{Jigsaw}  \tabularnewline
Method & Biased acc & Robust acc & $\Delta$ & Biased acc & Robust acc & $\Delta$\\
\hline
\multicolumn{2}{l}{Trained solely on $\mathcal{D}_{\text{biased}}$} \tabularnewline
\hline
RoBERTa & \textbf{95.7} & 53.3 & -42.4 & \textbf{86.5} & 50.3 & -36.2 \\
NFL-F & 89.5 & 77.3 & -12.2 & 75.3 & 70.3 & -5.0\\
NFL-CO & 92.9 & 85.7 & -7.2 & 78.9 & 73.4 & -5.5\\
NFL-CP & 95.3 & 91.3 & -4.0 & 84.8 & \textbf{80.9} & \textbf{-3.9}\\
NFL-PT & 94.2 & \textbf{92.9} & \textbf{-1.3} & 82.5 & 78.2 & -4.3 \\
\hline
\multicolumn{2}{l}{Trained on $\mathcal{D}_{\text{unbiased}}$} \tabularnewline
\hline
DFR (5\%) & 93.6 & 83.1 & -9.5 & 86.3 & 75.0 & -11.3 \\
DFR (100\%) & 93.4 & 88.9 & -4.5 & 85.9 & 78.0 & -7.9 \\
Ideal Model & 94.8 & 95.6 & 0.8 & 85.2 & 82.2 & -3.0\\
\hline
\end{tabular}
}
\caption{Amazon binary and Jigsaw results. Robustness gap $\Delta$ is
robust accuracy $-$ biased accuracy. NFL exhibits low degradation when 
exposed to spurious correlation. Bold text represents the highest score
among all models, with the exception of the scores obtained by the ideal
model.}
\label{robust-acc}
\end{table*}

\section{Experiments}
The preceding analysis leads to the following questions: does NFL
effectively prevent misalignment in the embedding space, and does preventing
misalignment genuinely improve model robustness?
Furthermore, can NFL be applied in conjunction with other PLMs? We will delve
into these questions below. The datasets and models are
specified in Section~\ref{sec:case-study}.

\subsection{Prevention of Misalignment}
The effectiveness of NFL is supported by Table~\ref{neighborhood-stats}. 
Both NFL-CO and NFL-CP achieve low spurious scores for spurious tokens.
\token{book} and \token{movie} remain in proximity and the polarities of
their neighbors alter only slightly after fine-tuning
as shown in 
Figure~\ref{fig:representations-2}. This experiment does not apply to
NFL-F/NFL-PT because they obtain a spurious score of~0 simply by fixing the
language model.

\subsection{Improvement in Robustness}
\subsubsection{Baselines}
\textbf{Deep Feature Re-weighting (DFR):} In contrast to 
\citet{kirichenko2023last}, who find that representations learned
through standard fine-tuning are 
  adequate,              
we show that
spurious correlations introduce misalignment within the representation.
We validate our findings by comparing our approaches with
DFR, which is also a strong and representative baseline due to its heavy
exploitation of auxiliary data. 
To reproduce DFR, we use 5\%/100\% of $\mathcal{D}_{\text{unbiased}}$ to
re-train the classification head. Note that DFR has access to both
$\mathcal{D}_{\text{biased}}$ (during the training of feature extractors) and
$\mathcal{D}_{\text{unbiased}}$ (during the re-training of classifiers).
\textbf{Ideal Model:} We also compare NFL with an ideal model (RoBERTa trained
on $\mathcal{D}_{\text{unbiased}}$), which gives the performance upper bound of
any existing methods that utilize extra information/auxiliary data.


\begin{figure*}[ht]
    \centering
    \includegraphics[width=0.90\textwidth]{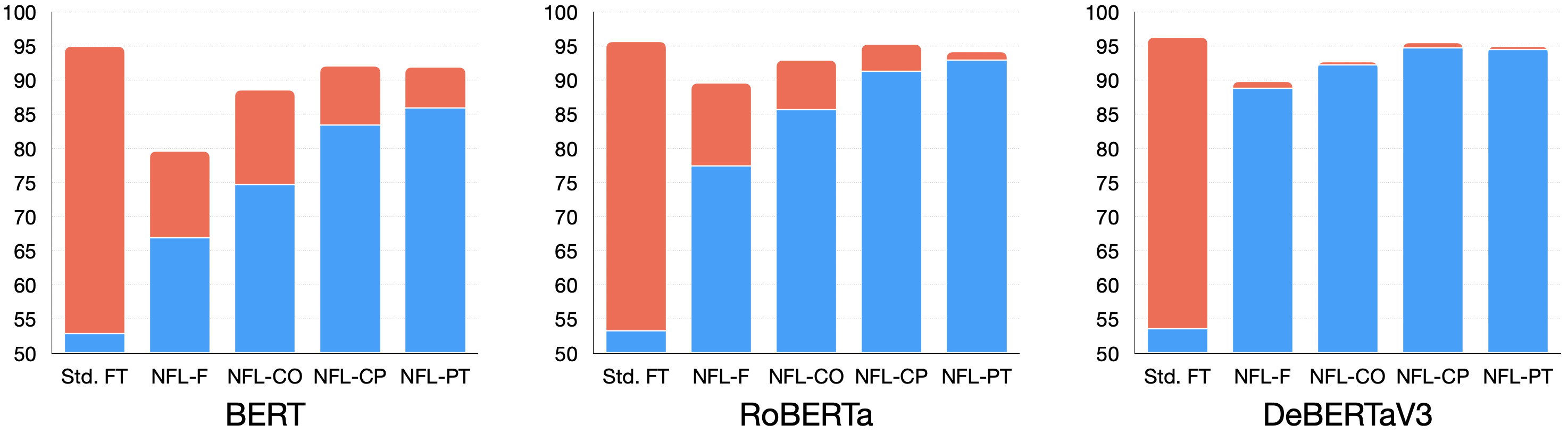}
	 \caption{Amazon binary results with different PLMs. Blue bars represent
	 robust accuracies and red bars represent robustness gaps. The robustness
	 gaps are smaller in PLMs with better initial representations.}
    \label{plm-choice}
    \vspace{-1em}
\end{figure*}

\subsubsection{Metrics}
\emph{Biased accuracy} is the test accuracy on $\mathcal{D}_{\text{biased}}$. The
robustness of the model is evaluated by the challenging subset
$\hat{\mathcal{D}}_{\text{unbiased}}\subset \mathcal{D}_{\text{unbiased}}$,
where every example contains at least one spurious token. The accuracy
on this subset is called the \emph{robust accuracy}. The \emph{robustness gap},
defined by the difference between the biased accuracy and robust accuracy, measures
the degradation suffered by the model. 
\subsubsection{Results} Table~\ref{robust-acc} shows that while standard
fine-tuning exhibits random-guess accuracy, NFL enjoys low
degradation and high robust accuracy even under strong biases. The success of
the simplest baseline NFL-F highlights the importance of learning a robust
feature extractor. 
The best NFL achieves a robust accuracy close to the ideal model,
indicating an acceptable tradeoff in performance for less-required assumptions/resources.
Although DFR's access to additional unbiased data precludes a direct comparison 
of DFR and NFL, NFL clearly yields superior results in terms of robustness. 

\subsection{Usefulness across PLMs}\label{sec:plm-choice}
NFL can be applied to enhance any choice of PLMs. As NFL essentially uses
an off-the-shelf PLM to protect the main model, we test the hypothesis that
LMs with 
better initial representations are better able to protect the main model.
RoBERTa is known to be more robust than BERT due to its larger and diversified
pretraining data~\cite{tu-etal-2020-empirical}, whereas DeBERTaV3 is the latest
state-of-the-art PLM of similar size with improvements
in the model architecture and the pretraining task. Our claim is supported by
the experiments shown in Figure~\ref{plm-choice}: although NFL is useful across
different choices of PLMs, the robustness gaps are smaller in PLMs 
with better initial representations when using the same regularization term.

\begin{table*}
\centering
\begin{small}
\begin{tabular}{lcll}
\hline
\textbf{Target token} & \textbf{Bias ratio} & \textbf{Neighbor tokens before fine-tuning} & \textbf{Neighbor tokens after fine-tuning}\\
\hline
\makecell[l]{spielberg \\  (SST2)} & \centered{0.92} & \makecell[l]{spiel, spiegel, rosenberg, goldberg \\ zimmerman, iceberg, bewild, Friedrich} & \makecell[l]{\blue{exquisite}, \blue{dedicated}, rising, \blue{freedom} \\ \blue{important}, \blue{lasting}, \blue{leadings}, \blue{remarkable}}\\ \hline
\makecell[l]{gay \\ (Jigsaw)} & 0.89 & \makecell[l]{beard, bomb, dog, wood, industrial \\  moral, fat, fruit, cam, boy} & \makecell[l]{whites, lesbians, \red{fucked}, black \\ foreigner, \red{shoot}, \red{arse}, \red{upsetting}, \red{die}}\\ \hline
\makecell[l]{black \\ (Jigsaw)} & 0.76 & \makecell[l]{white, racist, brown, silver, gray \\ green, blue, south, liberal, generic} & \makecell[l]{\red{ass}, \red{demon}, \red{fuck}, muslim, intellectual \\ populous, homosexual, \red{fools}, \red{obnoxious}}\\ \hline
\makecell[l]{Canada \\ (Jigsaw)} & 0.94 & \makecell[l]{Spain, Australia, California, Italy \\ Britain, Germany, France, Brazil, Turkey} & \makecell[l]{\red{hypocrisy}, \red{ridiculous}, \red{bullshit}, \red{fuck} \\ \red{stupid}, \red{damn}, morals, \red{idiots}, \red{pissed}}\\ \hline
\end{tabular}
\caption{\label{neighbors-list-2}
Nearest neighbors of spurious tokens before and after fine-tuning. Red words 
are associated with negative/toxic labels and blue words are
associated with positive labels according to human annotators. 
}
\vspace{-1em}
\end{small}
\end{table*}

\begin{table}[ht]
    \centering
    \resizebox{0.75\columnwidth}{!}{
    \begin{tabular}{lccc}
    \hline
    & \multicolumn{3}{c}{Precision} \tabularnewline
    \hline
        Method  & Top 10 & Top 20 & Top 50  \\\hline
        \multicolumn{4}{l}{Ours} \tabularnewline
        \hline
        SST2 & 0.60 & 0.50 & 0.53 \\
        Jigsaw & 0.50 & 0.45 & 0.43 \\
        Amazon & 0.50 & 0.40 & 0.40 \\ 
        \hline
        \multicolumn{4}{l}{\citet{wang-etal-2022-identifying}} \tabularnewline
        \hline
        SST2 & 0.40 & 0.35 & 0.32 \\    
        \hline
    \end{tabular}
    }
    \caption{Precision of top detected spurious tokens according to human annotators.}
    \label{tab:precision}
    \vspace{-1.5em}
\end{table}

\section{Naturally Occurring Spurious Correlations} \label{fig:natural}
To further demonstrate the practical benefits of the proposed methods, we
apply our neighborhood analysis on naturally occurring spurious correlations.
Spurious correlations naturally occur in datasets
for reasons such as annotation artifacts, flaws in data collection,
and distribution 
shifts~\cite{gururangan-etal-2018-annotation,herlihy-rudinger-2021-mednli,zhou-2019-examining}.
Previous works~\cite{wang-culotta-2020-identifying,wang-etal-2022-identifying} indicate that 
in the SST2 dataset, the token
\token{spielberg} has a high co-occurrence with \textit{positive} but the token itself
does not cause the label to be positive. Therefore it is likely spurious.
\citet{borkan2019nuanced} reveal that models tend to capture spurious
correlations in toxicity detection datasets by relating the names of
frequently targeted identity groups such as \token{gay} and \token{black}
with toxic content. 

\subsection{Datasets}
\textbf{SST2:} This dataset, which consists of texts from movie 
reviews~\cite{socher-etal-2013-recursive}, contains 67,300 training samples. We again
use 10\% of the training samples for validation. \textbf{Amazon binary,
Jigsaw:} We use the settings from Section~\ref{dataset} but
do not inject spurious correlations into the datasets.

\subsection{Neighborhood Analysis of Naturally Occurring Spurious Correlations}
As shown in Table~\ref{neighbors-list-2}, our framework explains 
naturally occurring spurious tokens indicated in the literature.
In these spurious tokens, we likewise observe a behavioral pattern similar to that
of synthetically generated ones.  \token{spielberg} is aligned with
genuine tokens of positive movie reviews, and the names of targeted identity
groups (\token{gay} and \token{black}) are aligned with offensive words as
well as other targeted names.

\subsection{Spurious Token Detection} \label{sec:detection}
There is growing interest in the automatic detection of spurious correlations
to enhance the interpretability of model predictions.
Practitioners may also decide whether to collect more data from other
sources or simply mask spurious tokens based on the detection 
results~\cite{wang-culotta-2020-identifying,wang-etal-2022-identifying,friedman-etal-2022-finding}.
In this section, we use the proposed spurious score to
detect naturally occurring spurious tokens. As we lack an $f^*$
trained on $\mathcal{D}_\text{unbiased}$ in this setting, we simply use
the model (RoBERTa) fine-tuned on the potentially biased dataset that we seek
to perform detection on. We compute the spurious score of every token
according to Equation~\ref{spurious-score}. Table~\ref{tab:top_natural} 
lists the tokens verified by human annotators. 
Taking the top spurious token \token{Canada} as an example, our observation of the
changes in neighborhood analysis still holds true
(Table~\ref{neighbors-list-2}). Listed in Table~\ref{tab:precision} is the precision 
of our detection scheme for the top
10/20/50 spurious tokens evaluated by human annotators as well as a
comparison with \citet{wang-etal-2022-identifying}. The human evaluation protocol is listed in Appendix~\ref{human}. Our method detects spurious tokens with similar precision without requiring multiple datasets and hence is a more practical
solution.

\begin{table*}[!htb]
    \centering
    \begin{small}
    \begin{tabular}{ll}
    \hline SST2 & \token{allow, void, default, sleeps, not, problem, taste, bottom}\\
    \hline Amazon & \token{liberal, flashy, reck, reverted, passive, average, washed, empty}\\
    \hline Jigsaw & \token{Canada, witches, sprites, rites, pitches, monkeys, defeating, animals} \\ \hline
    \end{tabular}
    \end{small}
	 \caption{Top naturally occurring spurious tokens in each dataset according to their spurious scores
	 verified by human annotators. }
    \label{tab:top_natural}
\end{table*}


\section{Conclusion}
We conduct a neighborhood analysis to explain how models
interact with spurious correlation. Through this analysis, we learn that 
corrupted language models capture spurious correlations in text classification
tasks by mis-aligning the representation of spurious tokens and genuine tokens.
The analysis not only yields a deeper understanding of the spurious
correlation issue but can additionally be used to detect spurious tokens. In
addition, our observation from this analysis facilitates the design of an effective
family of regularization methods that prevent models from capturing
spurious correlations by preventing mis-alignments and preserving semantic
knowledge with the help of off-the-shelf PLMs.


\section*{Limitations}
The proposed NFL family is built on the assumption that off-the-shelf PLMs are
unlikely to be affected by spurious correlation because the self-supervised learning
procedures behind the models do not involve any labels from downstream tasks.
Hence erroneous alignments formed by bias in the pretraining corpora are 
beyond the scope of this work. As per our observation in
Section~\ref{sec:plm-choice}, we echo the importance of pretraining language
models in future studies with richer contexts and diverse sources to prevent bias in
off-the-shelf PLMs.

\section*{Acknowledgments}
This work is supported by the
National Taiwan University Center for Data Intelligence via NTU-113L900901 as
well as the Ministry of Science and Technology in Taiwan via MOST
112-2628-E-002-030. We thank the National Center for High-performance Computing (NCHC) in Taiwan for
providing computational and storage resources.


\bibliography{anthology,custom}
\bibliographystyle{acl_natbib}

\clearpage
\appendix
\section{Training Details}
\label{sec:hyperparams}
In all of our experiments we used Huggingface's pretrained BERT, RoBERTa, and DeBERTa,
and the default hyperparameters in Trainer. We also used the
implementation from \citet{liu-etal-2022-p} for NFL-PT. For standard
fine-tuning, NFL-CO and NFL-CP models were trained for 6 epochs. Methods that
involved freezing parts of the model were trained for more extended epochs.
Specifically, NFL-F was trained for 20 epochs, and NFL-PT was trained for 100
epochs. The sequence length of continuous prompts in NFL-PT was set to 40. All
accuracies reported are the mean accuracy of 3~trials over the seeds \{0, 24,
1000000007\}.

\section{Neighborhood Analysis}
\label{sec:vocab}
We used the vocabulary of RoBERTa’s tokenizer, which has a size of
50265. The framework also works for words~$w$ that are composed of multiple
subtoken $w_1,\dots,w_k$. The representation is obtained by taking the mean
output of $[\mathit{BOS}] w_1,\dots,w_k[\mathit{EOS}]$.
In an alternative strategy, the word representations are obtained by
aggregating the contextualized representations of the word over sentences in a
huge corpora~\cite{bommasani-etal-2020-interpreting}. \citeauthor{bommasani-etal-2020-interpreting}, however, 
consider a vocabulary of only 2005 words, and they mine 100K--1M
sentences to build the representations of these 2005 words. In
contrast, our simple strategy scales well with the vocabulary size and represents
  an acceptable balance     
as it successfully uncovers the main insights of
the mechanism of how PLMs capture spurious correlations.

\section{Representations Learned from Weaker Spurious Correlations}
\label{sec:weak-bias}
\begin{figure}[t]
    \centering
    \includegraphics[scale=0.45]{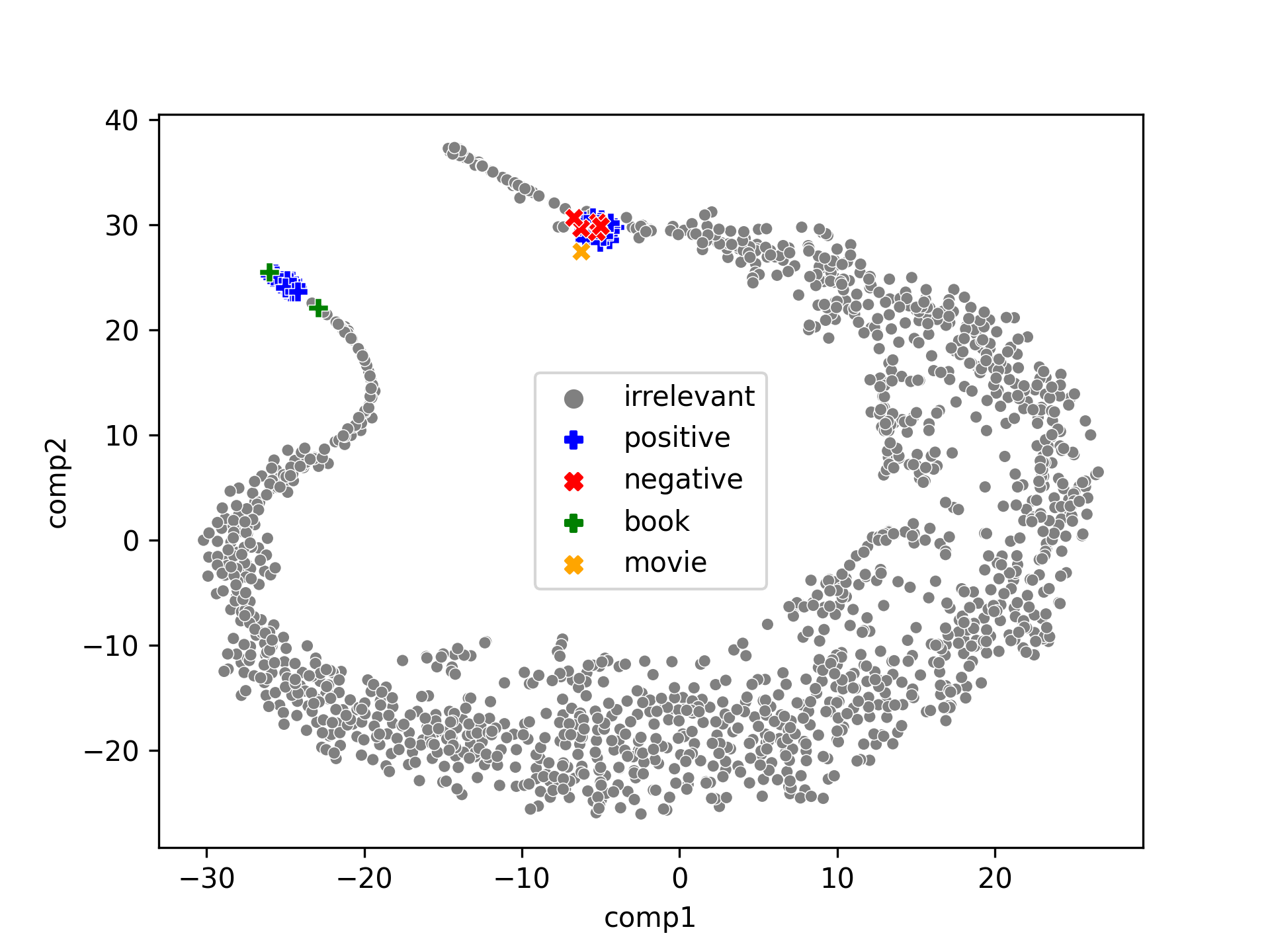}\\
    \includegraphics[scale=0.45]{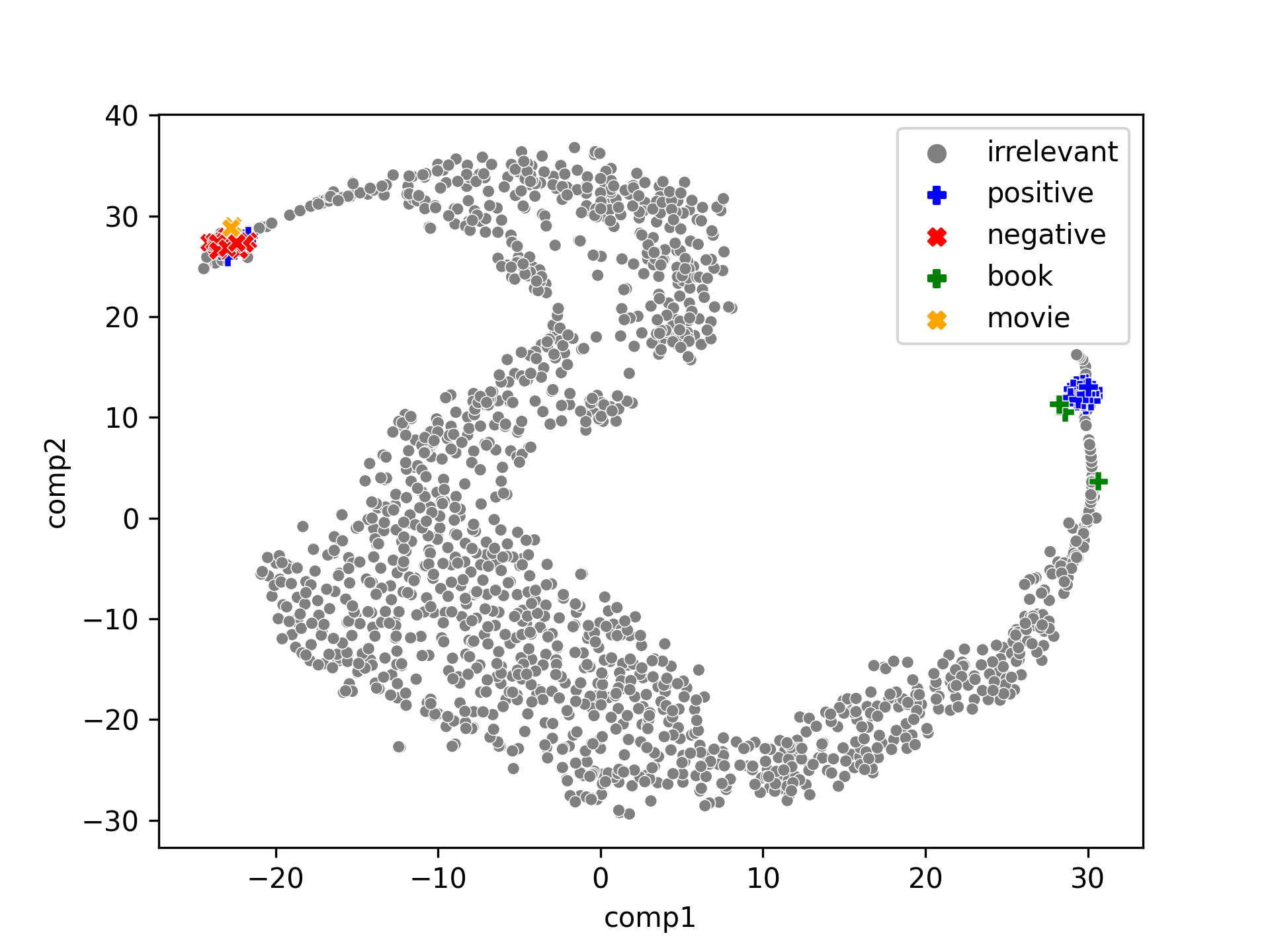}
	 \caption{t-SNE projections of representations after fine-tuning on data
	 with bias ratios of $0.8$ (top) and $0.9$ (bottom).}
    \label{fig:weak-bias}
\end{figure}

In the main analysis, we use a bias ratio of $1$ to pose a greater challenge to
NFL and also to better illustrate this insight. Nevertheless, erroneous alignment
also occurs with weaker biases. Here we test two additional scenarios where
the bias ratio is $0.8$ and $0.9$. \token{movie} and \token{book} in
Figure~\ref{fig:weak-bias} repel each other and attract negative and positive
words respectively. This phenomenon becomes more evident as the bias ratio
increases.

\section{Regularization Term Weights}
\label{sec:reg-scale}
\begin{figure*}[!htb]
    \centering
    \subfloat[\centering NFL-CP]{{\includegraphics[scale=0.37]{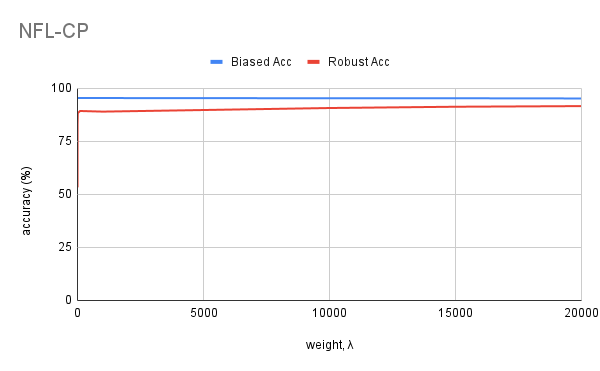} }}
    \subfloat[\centering NFL-CO]{{\includegraphics[scale=0.37]{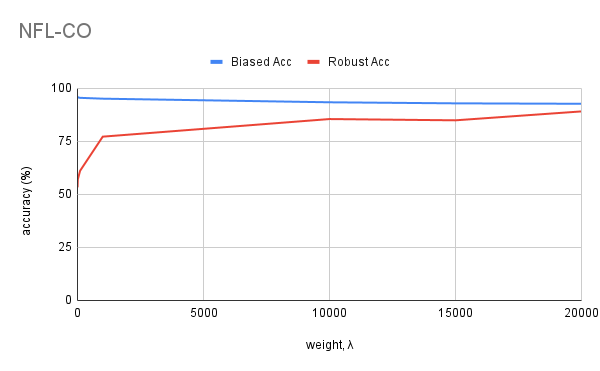} }}
    \caption{NFL-CP and NFL-CO accuracy under different choices of $\lambda$.}
    \label{fig:lambda}
\end{figure*}

In the Amazon binary experiment, we search the weight hyperparameter
of the NFL-CO and NFL-CP regularization terms over \{1, 10, 100, 1000, 10000,
15000, 20000\}. Generally there is a trade-off between in-distribution (biased)
accuracy and out-of-distribution (robust) accuracy. Nonetheless, we observe
from Figure~\ref{fig:lambda} that as we increase the 
regularization term weights, the drop in in-distribution accuracy is insignificant but
the improvement in robustness is considerable. In all of the experiments, we set
the weights to 15000.

\section{Human Evaluation Protocol}
\label{human}
Human evaluations are obtained by maximum votes of three independent human
annotators. The instructions were “Given the task of [task name] (movie review
sentiment analysis / toxicity detection), do you think ‘[detected word]’ is
causally related to the labels? Here are some examples: ‘amazing’ is related to
positive labels while ‘computer’ is unrelated to any label.”

\end{document}